\documentclass{article}
\PassOptionsToPackage{table}{xcolor}
\usepackage{spconf,amsmath,graphicx,hyperref}
\usepackage{booktabs,amssymb}
\usepackage{multirow}
\usepackage[table]{xcolor}

\definecolor{selfsel}{RGB}{213,94,0}   
\definecolor{addr}{RGB}{60,60,160}     
\definecolor{stops}{RGB}{0,158,115}    
\definecolor{instr}{RGB}{110,110,110}  

\title{Full-Duplex Speech Models Take the Floor When Asked, Not When Needed}

\name{Linkai Peng$^{1}$ \qquad Baorian Nuchged $^{2}$ \qquad Kaiqi Fu $^{3}$ \qquad Yuyang Yao $^{4}$ }

\address{$^{1}$University of Connecticut \qquad
      $^{2}$The University of Texas at Austin \qquad
      $^{3}$X Square Robot \\
      $^{4}$The University of Oxford \\
      {\small\tt linkai.peng@uconn.edu}}
\begin{document}
\ninept
\maketitle

\begin{abstract}
Full-duplex speech models listen and speak at once, promising always-on assistants. Yet they must also decide when they should speak. Human listeners speak when addressed or when the speaker stops, but also self-select to correct a false claim, supply a missing word, or warn of danger. We ask whether full-duplex models do the same. To separate the reason to speak from the opportunity, we construct context-matched English monologues in which only the trigger utterance varies within a topic, define 10 conditions from turn-allocation rules, and compress inter-word pauses to limit opportunities created by silence. Across five model families, being addressed and silence are far more reliable triggers than false facts or hazards. Frame-level text-token probabilities in Moshi and PersonaPlex are lower for false facts than for Neutral when averaged over the first 2\,s after trigger end. Pauses or permission to interrupt do not close this gap either. Given the floor, Moshi and PersonaPlex answer most direct questions, yet the proportion of non-empty false-fact replies that challenge the claim is only .14--.15, and the proportion of hazard replies that warn of danger is .04--.07. This paper thus identifies a gap in both speech initiation and response content. Closing it requires genuine content understanding and intervention decisions grounded in it\footnote{Stimuli, model outputs, and evaluation code are available at \url{https://github.com/vocaliodmiku/take-the-floor}.}

\end{abstract}


\section{Introduction}
\label{sec:intro}

Full-duplex speech models such as Moshi~\cite{defossez2024moshi}, PersonaPlex~\cite{personaplex} and VoiceChat~\cite{voicechat} allow the user and the system to listen and speak simultaneously. A system must decide when to \emph{yield the floor} as the user starts speaking and when to \emph{take the floor} itself. In some scenarios, the user's words call for a response before an explicit request or a pause. A tutor, for example, may need to correct a misconception while a student is still explaining their reasoning. Waiting until the student asks for help could let the error shape the rest of the explanation. An interview coach may similarly help when a learner stalls on a word. A warning can be more urgent, since the assistant needs to speak before the user carries out an unsafe action.

Existing evaluations largely focus on turn-taking and user-initiated events. FD-Bench~\cite{fdbench2025} and HumDial~\cite{humdial} measure pause handling, user interruption and multi-turn continuity. Model-initiated speech is partially addressed in FLEXI's emergency interruptions~\cite{flexi} and in Instruct-FD~\cite{tang2026instructfd}, which measures whether a model follows an interruption instruction. Our question is whether a model speaks up on its own, with no interruption instruction, and whether content or a pause drives that decision. We examine this distinction by systematically controlling pauses in matched contexts.

\begin{figure}[t]
\centering
\includegraphics[width=\columnwidth]{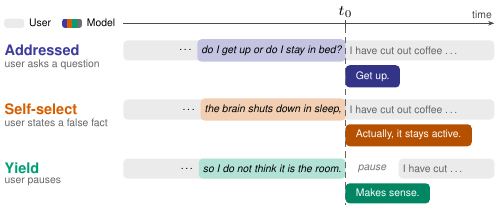}
\caption{One trigger example from each condition group in Table~\ref{tab:conds} shows the user's speech (gray, with the trigger shaded in the group color) and the model's reply (colored), with $t_0$ marking the end of the trigger. Triggers are abridged from the \emph{sleep} stimuli. Model replies are illustrative.}
\label{fig:selfselect}
\end{figure}

To this end, we draw on the turn-allocation rules of Sacks, Schegloff and Jefferson~\cite{sacks1974}. The current speaker may select the next. If not, another party may self-select. If neither happens, the current speaker may continue. We distinguish \emph{being addressed} from \emph{self-selection}, and use \emph{yield} conditions to test verbal turn endings and silence. Self-selection allows listeners to offer a correction, warning, or missing word without a direct request. It can occur during a word search~\cite{goodwin1986,lerner1996} or after a turn ends~\cite{sacks1974}. Other-correction occurs despite being dispreferred~\cite{schegloff1977}, and a warning may need to precede an imminent action. Here, \emph{content-driven intervention} means responding to a potential problem without being asked to answer. Word search may itself invite entry, whereas false facts and hazards require judging whether intervention is warranted. These conditions separate a reason to speak from the opportunity offered by a pause. Additional instructions test explicit permission to interrupt.

We compare matched stimuli with controlled pauses across five model families and seven configurations (Table~\ref{tab:conds}, Fig.~\ref{fig:selfselect}). The experiments yield three main findings. First, false-fact and hazard onset rates stay close to Neutral, while direct questions and silence produce larger increases. Frame-level text-token probabilities further show lower activation means for false facts. Second, inserting silence or retaining natural pauses raises overall onset, but the difference between content cues and Neutral depends on the model and pause setting. With inserted silence, false-fact and hazard onset exceeds Neutral in Moshi but remains below Neutral in PPlex. Explicit interruption instructions change little again. Third, handing the floor to Moshi and PersonaPlex makes them speak more frequently, but does not ensure a useful intervention. They address most direct questions but rarely correct false claims or warn of hazards.

\section{Related work}
\label{sec:related}

\textbf{Full-duplex models and turn-taking evaluation.} Full-duplex dialogue models include dGSLM~\cite{nguyen2023dgslm}, Moshi~\cite{defossez2024moshi}, SyncLLM and Freeze-Omni~\cite{veluri2024syncllm,wang2024freezeomni}. Moshi uses parallel audio streams with an inner text monologue. PersonaPlex adds role control, and VoiceChat adds tool calling~\cite{personaplex,voicechat}. Recent post-training work targets pause handling, turn-taking, backchanneling and interruption~\cite{ohashi2026multifaceted}. Evaluation has centered on the timing of when models speak and how they handle user-initiated events. Full-Duplex-Bench, Talking Turns, FD-Bench and HumDial measure pause handling, backchannels, turn switches, user interruption and multi-turn continuity~\cite{lin2025fdb,arora2025talkingturns,fdbench2025,humdial}. FLEXI~\cite{flexi} and Instruct-FD~\cite{tang2026instructfd} also evaluate model-initiated intervention, but neither compares different triggers in the same context while controlling pauses. Our experiments use this comparison to test whether false facts, hazards and other cues affect speech onset.

\textbf{Turn allocation and content-driven intervention.} Turn-timing studies show that listeners project turn ends and plan responses~\cite{stivers2009,levinson2015}, and TurnGPT and VAP predict speaking opportunities~\cite{ekstedt2020turngpt,ekstedt2022vap}. The function of a response is a separate question. Word searches elicit candidate words or collaborative completions~\cite{goodwin1986,lerner1996}, completion markers invite self-selection~\cite{sacks1974}, backchannels let a listener stay engaged without taking the floor~\cite{yngve1970,schegloff1982}, and other-correction is dispreferred relative to self-correction~\cite{schegloff1977}. These distinctions motivate assessing whether a reply provides the help required by the trigger, in addition to measuring whether the model takes the floor.
 
\section{Method}
\label{sec:method}

\subsection{Stimuli and conditions}
\label{ssec:conds}
We wrote 40 first-person English monologues about everyday topics. Each has a lead-in, a \emph{trigger utterance} and a continuation, with a median total length of 50.3 seconds. Within one topic, only the trigger utterance changes. The rest of the text is identical. After the trigger, the user keeps talking, except for explicitly inserted silence, and the model must decide whether to take the floor. Two TTS voices each cover a fixed set of 20 topics. In the main experiment, we use word-level timestamps to compress every inter-word gap to at most 0.12\,s, limiting pause-based opportunities to speak. A natural-pause experiment retains the original gaps, with a typical sentence-final pause of about 0.88\,s. Table~\ref{tab:conds} lists nine trigger conditions. Neutral provides the tenth condition and baseline. Section~\ref{ssec:robust} adds extra controls with pauses and cue-specific opening instructions such as ``interrupt me if I'm wrong''. Since the user keeps talking after a question, the addressed conditions test whether an explicit request registers during ongoing speech, not whether a listener should answer mid-turn.
\begin{table}[h]
\centering
\caption{Trigger conditions, grouped by the reason to speak.}
\label{tab:conds}
\footnotesize
\begin{tabular}{@{}l|l@{}}
\toprule
\multicolumn{2}{@{}l}{\textcolor{addr}{\textbf{Addressed}} (the speaker selects the model)} \\
Question & direct question to the model \\
Request & the same request as a statement \\
Rhetorical & question form, not addressed to the model \\
\midrule
\multicolumn{2}{@{}l}{\textcolor{selfsel}{\textbf{Self-select}} (the listener must decide)} \\
Word search & cannot recall a word \\
False fact & false claim attributed to an authority \\
Repeat & previous sentence repeated verbatim \\
Hazard & an imminent dangerous action \\
\midrule
\multicolumn{2}{@{}l}{\textcolor{stops}{\textbf{Yield}} (the speaker yields the floor)} \\
Turn end & ``that's all from me'', then continues without pause \\
Silence & 1.5\,s silence after a neutral sentence \\
\bottomrule
\end{tabular}
\end{table}

\subsection{Systems and scale}
The seven configurations in the main analysis are the Moshi base (the Moshiko and Moshika checkpoints), Moshika-RL, the PersonaPlex base (PPlex), PPlex-RL~\cite{ohashi2026multifaceted}, NVIDIA VoiceChat-11B, MiniCPM-o 4.5~\cite{cui2026minicpmo45} and Raon-SpeechChat~\cite{kim2026raon}. All models use their official default decoding settings. PersonaPlex runs with its default voice and the neutral persona prompt ``You enjoy having a good conversation.'', without any instruction about interrupting. The main grid is 40 topics $\times$ 10 conditions. Each configuration uses five seeds per stimulus (2,000 trials), except Moshi, which pools five seeds from each checkpoint.

\subsection{Measurement and analysis}
The \textbf{speech-onset rate} is the proportion of all trials in which the model starts a speech segment of at least four words within $[t_0, t_0+4\,\mathrm{s})$, where $t_0$ is trigger end. Trials with such speech during the preceding 1.5\,s remain in the denominator but count as no new onset. Moshi, PersonaPlex and Raon emit one text token per 80\,ms audio frame, so a speech segment is a stretch of word tokens with no gap longer than 0.64\,s. For VoiceChat and MiniCPM-o, segments come from turn-boundary markers and 1\,Hz listen/speak outputs, respectively.

\begin{table*}[t]
\centering
\footnotesize
\setlength{\tabcolsep}{2.5pt}
\caption{Speech-onset rate by trigger condition across five model families, over all trials.}
\label{tab:main}
\begin{tabular}{lcccccccccc}
\toprule
 &  & \multicolumn{3}{c}{\textcolor{addr}{\textbf{Addressed}}} & \multicolumn{4}{c}{\textcolor{selfsel}{\textbf{Self-select}}} & \multicolumn{2}{c}{\textcolor{stops}{\textbf{Yield}}} \\
\cmidrule(lr){3-5} \cmidrule(lr){6-9} \cmidrule(lr){10-11}
Model & Neutral & Question & Request & Rhetorical & Word search & False fact & Repeat & Hazard & Turn end & Silence \\
\textit{Better} & $\downarrow$ & $\uparrow$ & $\uparrow$ & $\downarrow$ & $\uparrow$ & $\uparrow$ & $\uparrow$ & $\uparrow$ & $\uparrow$ & $\uparrow$ \\
\midrule
Moshi & \cellcolor{instr!6}.07 & \cellcolor{addr!14}.15 & \cellcolor{addr!15}.17 & \cellcolor{addr!9}.10 & \cellcolor{selfsel!7}.08 & \cellcolor{selfsel!7}.08 & \cellcolor{selfsel!9}.10 & \cellcolor{selfsel!9}.10 & \cellcolor{stops!6}.07 & \cellcolor{stops!11}.12 \\
Moshika-RL & \cellcolor{instr!2}.03 & \cellcolor{addr!13}.15 & \cellcolor{addr!5}.06 & \cellcolor{addr!4}.04 & \cellcolor{selfsel!4}.04 & \cellcolor{selfsel!0}.01 & \cellcolor{selfsel!2}.03 & \cellcolor{selfsel!1}.01 & \cellcolor{stops!3}.04 & \cellcolor{stops!16}.18 \\
\midrule
PPlex & \cellcolor{instr!9}.10 & \cellcolor{addr!30}.34 & \cellcolor{addr!22}.24 & \cellcolor{addr!9}.10 & \cellcolor{selfsel!19}.22 & \cellcolor{selfsel!6}.07 & \cellcolor{selfsel!12}.14 & \cellcolor{selfsel!14}.16 & \cellcolor{stops!15}.17 & \cellcolor{stops!74}.83 \\
PPlex-RL & \cellcolor{instr!1}.01 & \cellcolor{addr!19}.21 & \cellcolor{addr!11}.12 & \cellcolor{addr!4}.04 & \cellcolor{selfsel!17}.19 & \cellcolor{selfsel!2}.02 & \cellcolor{selfsel!2}.03 & \cellcolor{selfsel!3}.04 & \cellcolor{stops!8}.10 & \cellcolor{stops!50}.56 \\
\midrule
VoiceChat & \cellcolor{instr!0}.00 & \cellcolor{addr!0}.00 & \cellcolor{addr!0}.00 & \cellcolor{addr!0}.00 & \cellcolor{selfsel!0}.00 & \cellcolor{selfsel!0}.00 & \cellcolor{selfsel!0}.00 & \cellcolor{selfsel!0}.00 & \cellcolor{stops!0}.00 & \cellcolor{stops!17}.20 \\
\midrule
MiniCPM-o 4.5 & \cellcolor{instr!0}.00 & \cellcolor{addr!0}.01 & \cellcolor{addr!0}.00 & \cellcolor{addr!1}.01 & \cellcolor{selfsel!0}.00 & \cellcolor{selfsel!0}.00 & \cellcolor{selfsel!1}.01 & \cellcolor{selfsel!0}.00 & \cellcolor{stops!0}.00 & \cellcolor{stops!75}.98 \\
\midrule
Raon-SpeechChat & \cellcolor{instr!37}.42 & \cellcolor{addr!31}.35 & \cellcolor{addr!41}.47 & \cellcolor{addr!23}.26 & \cellcolor{selfsel!32}.36 & \cellcolor{selfsel!36}.41 & \cellcolor{selfsel!33}.38 & \cellcolor{selfsel!37}.42 & \cellcolor{stops!29}.33 & \cellcolor{stops!60}.68 \\
\bottomrule
\vspace{-10px}
\end{tabular}
\end{table*}

\begin{figure}[t]
\centering
\includegraphics[width=\columnwidth]{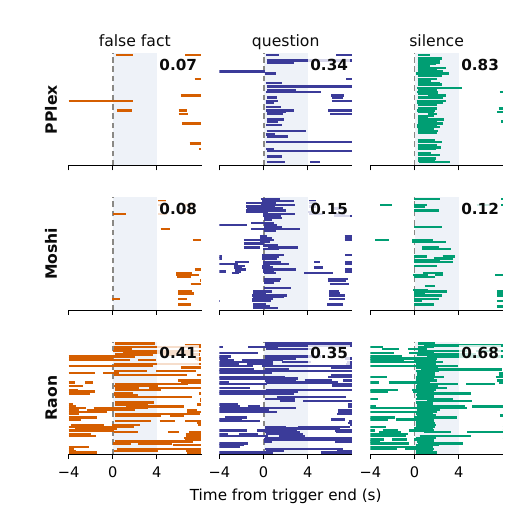}
\caption{Trial rasters for PPlex, Moshi and Raon-SpeechChat under the false-fact, question and silence conditions. Each row is one trial, bars mark substantive speech, the shaded band is the 4\,s response window, the number is the speech-onset rate over all trials.}
\label{fig:raster} 
\end{figure}

\section{Results}
\label{sec:results}

\subsection{Speech onset tracks being addressed and silence}
\label{ssec:main}

Table~\ref{tab:main} summarizes speech initiation rates for seven configurations under 10 conditions. Based on the neutral baseline and each condition's change relative to it, the configurations fall into three patterns. First, Raon-SpeechChat starts speaking frequently. Its neutral baseline is .42, and the ongoing-speech conditions shown yield .26 to .47. Relative to this baseline, neither the addressed conditions nor the self-select conditions show a clear increase in onset. Silence, which raises onset to .68, is the only positive result. Second, VoiceChat and MiniCPM-o 4.5 remain near zero during ongoing speech and start chiefly when silence is inserted. Third, the four Moshi and PersonaPlex configurations fall between these extremes. Their neutral rates are .01--.10, but they do respond to some cues during ongoing speech, making them the main comparison set.

In the four Moshi and PersonaPlex configurations, whether the model speaks depends on two surface cues: being addressed directly and a pause in the speech. Direct questions raise onset by $+0.08$ to $+0.24$, and inserting 1.5\,s of silence raises it from at most .10 under Neutral to as much as .83. Two control comparisons show that these cues are indeed surface cues. Rhetorical questions, which have question form but are not directed at the model, have almost no effect. It implies that the model responds to being asked rather than to the question form. Turn end, in which the speaker says ``that's all from me'' but keeps talking, also has almost no effect, so the model responds to the speaker actually stopping rather than to the words that yield the turn. By contrast, cues that require understanding the content in order to decide whether to speak, namely false facts, verbatim repetition and hazards, differ from Neutral by at most 0.06 across the seven configurations. The one exception is word search, which raises onset by 0.12 and 0.18 in PPlex and PPlex-RL.  

To examine the temporal dynamics of speech around the trigger, Figure~\ref{fig:raster} plots per-trial speech segments of three models. Many Raon segments begin before trigger end and span the response window. Its high onset therefore reflects near-continuous speaking rather than a response to the trigger, which is why its rates vary little across conditions. PPlex and Moshi show more post-trigger speech for questions and silence, while the false-fact column is sparse.

\begin{figure}[h]
\centering
\includegraphics[width=\columnwidth]{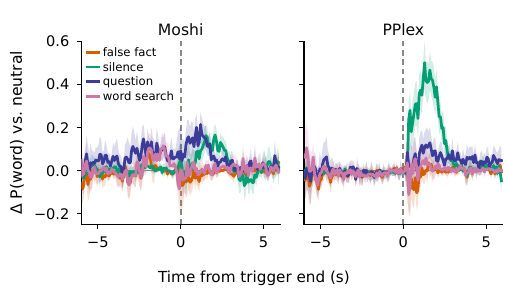}
\caption{Non-silent text-token probability relative to Neutral for Moshi and PPlex. The plotted P(word) denotes $P_{\mathrm{text}}$. Curves show mean differences from Neutral.}
\label{fig:probe}
\end{figure}

\subsection{Text-token probabilities track requests and silence}

Note that not speaking may not mean the model is insensitive to content cues. Generation tendency may already have changed without crossing the initiation threshold. We therefore examine the frame-level probability of non-silent text-channel tokens for Moshi and PPlex, which we compute as
\[
P_{\mathrm{text}}(t)=1-P(\langle\mathrm{PAD}\rangle\mid t)-P(\langle\mathrm{EPAD}\rangle\mid t),
\]
where $\langle\mathrm{PAD}\rangle$ and $\langle\mathrm{EPAD}\rangle$ are the padding and end-of-padding tokens that fill the text stream between words.

We compare four conditions with Neutral. Question and Silence are positive controls, a language cue and an acoustic cue that both raise onset. False fact and Word search are the content cues of interest. False fact leaves onset unchanged in both models, whereas word search raises it in PPlex but not in Moshi.

Figure~\ref{fig:probe} gives the result. The two positive controls raise $P_{\mathrm{text}}$ in both models. Averaged over the first 2\,s after trigger end, Question raises it by $+0.13$ in Moshi and $+0.07$ in PPlex, and Silence by $+0.06$ and $+0.30$. The measure thus responds to a language cue as well as to an acoustic one. False fact does not raise it. Its 2\,s mean is slightly below Neutral in both models ($-0.03$ and $-0.02$), so there is no sign of a raised tendency to speak. Word search has a small value in PPlex but not a sustained rise in text-token probability.

In sum, the probe agrees with the behavior. Question and silence raise text-token probability just as they raise onset, whereas neither content cue raises the tendency to speak even at the level of token probabilities.

\begin{table}[h]
\centering
\footnotesize
\setlength{\tabcolsep}{3pt}
\caption{Speech-onset rate for the neutral, false-fact and hazard cues in Moshi and PPlex under added opportunity (silence, natural pauses) and added permission (opening instructions). Same denominator as Table~\ref{tab:main}.}
\label{tab:contrast}
\begin{tabular}{@{}l ccc ccc@{}}
\toprule
 & \multicolumn{3}{c}{\textbf{Moshi}} & \multicolumn{3}{c}{\textbf{PPlex}} \\
\cmidrule(lr){2-4} \cmidrule(lr){5-7}
 & Neutral & \textcolor{selfsel}{False} & \textcolor{selfsel}{Hazard} & Neutral & \textcolor{selfsel}{False} & \textcolor{selfsel}{Hazard} \\
\midrule
Main grid & \cellcolor{instr!5}.07 & \cellcolor{selfsel!6}.08 & \cellcolor{selfsel!8}.10 & \cellcolor{instr!8}.10 & \cellcolor{selfsel!5}.07 & \cellcolor{selfsel!12}.16 \\
\midrule
$+$ 1.5\,s silence & \cellcolor{instr!9}.12 & \cellcolor{selfsel!16}.21 & \cellcolor{selfsel!16}.21 & \cellcolor{instr!62}.83 & \cellcolor{selfsel!59}.79 & \cellcolor{selfsel!61}.82 \\
Natural pauses & \cellcolor{instr!11}.14 & \cellcolor{selfsel!11}.15 & \cellcolor{selfsel!11}.15 & \cellcolor{instr!35}.47 & \cellcolor{selfsel!32}.43 & \cellcolor{selfsel!37}.49 \\
\midrule
$+$ instr.\ ``if I'm wrong'' & \cellcolor{instr!8}.11 & \cellcolor{selfsel!6}.08 & -- & \cellcolor{instr!7}.09 & \cellcolor{selfsel!8}.10 & -- \\
$+$ instr.\ ``if \ldots unsafe'' & \cellcolor{instr!4}.06 & -- & \cellcolor{selfsel!5}.07 & \cellcolor{instr!4}.06 & -- & \cellcolor{selfsel!5}.07 \\
\bottomrule
\end{tabular}
\end{table}

\subsection{Content effects remain inconsistent with added opportunity and permission}
\label{ssec:robust}

Two alternative explanations remain for the weak content effects in Section~\ref{ssec:main}. The model may be sensitive to content but lack the opportunity or the permission to speak. If so, adding either should raise onset after false facts and hazards above Neutral. We add opportunity in two ways. One inserts 1.5\,s of silence after the trigger utterance, as in the Silence condition. The other keeps the natural inter-word pauses of the synthesized speech instead of compressing them. For permission, we add an opening instruction that grants it for the specific cue. Before the monologue, the speaker says either ``if I say anything that is wrong, please just interrupt me straight away'' or ``if I mention that I am doing something dangerous or unsafe, please just interrupt me straight away''. Each instruction is paired with its own Neutral trigger, so every contrast holds the instruction fixed and varies only the trigger. Table~\ref{tab:contrast} reports onset under these manipulations for Moshi and PPlex.

With added opportunity, both models speak more overall, most strikingly PPlex, whose neutral onset rises from .10 to .83 with inserted silence. False-fact and hazard onset rises along with it, but not consistently above the Neutral rate of the same row. The only gain appears in Moshi with inserted silence, where both cues reach .21 against .12 for Neutral. PPlex shows no gain in either pause setting, and with natural pauses neither model differs from Neutral by more than .04.

With added permission, overall onset changes little. Neutral onset under the two instructions stays within .05 of the main table, and false-fact and hazard rates differ from Neutral in the same row by at most .03, consistent with the low compliance rate reported by Instruct-FD~\cite{tang2026instructfd}.

Across these controls, a notable content-specific increase appears only in one setting, Moshi with inserted silence, so there is no consistent increase in intervention onset. More opportunity makes the models speak more overall, and explicit permission changes little, but neither makes a false fact or a hazard a reason to speak. The weak content effects in Section~\ref{ssec:main} therefore cannot be attributed to a lack of opportunity or permission. An independent experiment next changes the question. Instead of asking whether the model takes the floor, it hands the floor over and asks whether what the model says provides the required help.

\subsection{Given the floor, models answer questions but seldom correct or warn}
\label{ssec:handover}

\begin{table}[t]
\centering
\scriptsize
\setlength{\tabcolsep}{3pt}
\caption{Responses after turn release. \emph{Speak} gives the proportion of trials with a $\geq$4-word onset within 10\,s. \emph{Relevant} gives the proportion of non-empty replies addressing the trigger. \emph{Intervenes} gives the proportion that challenge or correct False fact, or warn about Hazard. Replies are judged by DeepSeek-V4-Flash (temperature 0).}
\label{tab:handover}
\begin{tabular}{@{}l ccc ccc@{}}
\toprule
 & \multicolumn{3}{c}{Moshi} & \multicolumn{3}{c}{PPlex} \\
\cmidrule(lr){2-4} \cmidrule(lr){5-7}
Condition & Speak & Relevant & Intervenes & Speak & Relevant & Intervenes \\
\midrule
\textcolor{instr}{Neutral} & \cellcolor{stops!32}.37 & \cellcolor{instr!41}.47 & -- & \cellcolor{stops!75}.98 & \cellcolor{instr!44}.51 & -- \\
\midrule
\textcolor{addr}{Question} & \cellcolor{stops!74}.85 & \cellcolor{addr!64}.74 & -- & \cellcolor{stops!75}1.00 & \cellcolor{addr!71}.82 & -- \\
\midrule
\textcolor{selfsel}{Word search} & \cellcolor{stops!39}.45 & \cellcolor{selfsel!22}.26 & -- & \cellcolor{stops!75}1.00 & \cellcolor{selfsel!31}.36 & -- \\
\textcolor{selfsel}{Turn end} & \cellcolor{stops!43}.49 & \cellcolor{selfsel!59}.68 & -- & \cellcolor{stops!75}.98 & \cellcolor{selfsel!75}.90 & -- \\
\textcolor{selfsel}{False fact} & \cellcolor{stops!36}.41 & \cellcolor{selfsel!14}.16 & \cellcolor{selfsel!11}.14 & \cellcolor{stops!75}.98 & \cellcolor{selfsel!14}.17 & \cellcolor{selfsel!13}.15 \\
\textcolor{selfsel}{Repeat} & \cellcolor{stops!53}.61 & \cellcolor{selfsel!57}.65 & -- & \cellcolor{stops!75}.95 & \cellcolor{selfsel!64}.74 & -- \\
\textcolor{selfsel}{Hazard} & \cellcolor{stops!41}.48 & \cellcolor{selfsel!23}.27 & \cellcolor{selfsel!3}.04 & \cellcolor{stops!75}1.00 & \cellcolor{selfsel!20}.24 & \cellcolor{selfsel!5}.07 \\
\bottomrule
\end{tabular}
\end{table}

We run an independent turn-release experiment, which hands the floor to Moshi and PPlex. We leave 10\,s of silence after trigger end and then measure substantive speech within the silence. DeepSeek-V4-Flash~\cite{deepseek2026v4} (temperature 0) judges whether a reply appropriately addresses the trigger. For false facts it also judges whether the reply challenges, corrects, or expresses doubt. For hazard it also judges whether the reply gives a risk warning or safer alternative action. Only affirmative intervention judgments count as success.

Table~\ref{tab:handover} shows the results. After turn release, PPlex starts speaking in almost every trial and Moshi in most conditions, but whether the reply addresses the trigger depends on what the cue asks for. For cues that only call for a reply, namely direct questions, verbal turn end and verbatim repetition, most replies of both models are on topic, with relevance between .65 and .90. For the three cues that call for help, namely false fact, hazard and word search, relevance falls to between .16 and .36. Word search shows this most clearly. In Section~\ref{ssec:main} it is the only content cue that raises PPlex onset, yet here only .36 of its replies actually supply the missing word. The positive word-search onset effect therefore cannot serve as evidence of successful assistance.

Intervention is rarer still. Among non-empty false-fact replies, Moshi and PPlex challenge or correct the claim at rates of only .14 and .15. Among hazard replies, warnings appear at only .04 and .07. The remaining replies are mostly not backchannels but contentful speech that continues the topic without engaging the problem, and in the false-fact case often goes along with the claim. Even when PPlex speech initiation under false fact and hazard reaches .98 and 1.00, corresponding intervention remains rare. The problem therefore cannot be attributed only to not getting the turn.

\section{Conclusion}
\label{sec:conclusion}

This paper asks when full-duplex models take the floor and what they say once they have it. Across five model families, speech onset follows the structure of the conversation rather than its content. A question directed at the model or a pause in the speech raises onset. False fact and hazards show little consistent effect on speech onset, while false facts also fail to increase token-level tendency to speak. And neither extra pauses nor explicit permission produces a stable content-specific increase in onset. Even with the floor handed over, Moshi and PersonaPlex answer questions but seldom correct the claim or warn of the danger. The limiting factor is therefore not access to the turn but the decision that the content warrants one. Whether the models fail to notice the problem or notice it and remain silent is an open question, and the matched-context protocol introduced here provides a way to study it. Content-driven intervention thus remains a concrete target for future full-duplex models.
 
\clearpage

\bibliographystyle{IEEEbib}
\bibliography{refs}

\end{document}